\documentstyle[fleqn,leqno,eaclap]{article}

\newcommand{\avm}[1]{{ \setlength{\arraycolsep}{1mm}
                       \renewcommand{\arraystretch}{1}      
                       \hspace*{-0.15em} \left[
                       \begin{array}{@{}l@{~~}l@{}}
                         \\[-2.8mm] #1 \\[-2.8mm]           
                       \end{array}
                       \right] \hspace*{-0.15em} }}

\newcommand{\typedavm}[2]{{\begin{array}{@{}l@{}}
                           {\mbox{\scriptsize \it #1 }}^{\avm{#2}}
                          \end{array}}}

\newcommand{\attval}[2]{\mbox{\attribute{#1}} & #2 \\}

\newcommand{\attribute}[1]{{\sc #1}}

\title{Stochastic HPSG}
\author{Chris Brew \\
Language Technology Group \\
HCRC, University of Edinburgh \\
2 Buccleuch Place \\  Edinburgh EH8 9LW \\ Scotland, UK \\
email: {\tt Chris.Brew@edinburgh.ac.uk}\\}
\begin{document}
\bibliographystyle{acl}
\maketitle
\begin{abstract}
In this paper we
provide a
probabilistic interpretation for
typed feature structures very similar to those
used by Pollard and Sag.
We begin with a version of the interpretation which
lacks a treatment of re-entrant feature structures,
then provide an extended interpretation which
allows them.
We sketch algorithms allowing the numerical parameters of
our probabilistic interpretations of HPSG  to be estimated
{} from corpora.
\end{abstract}
\section{Introduction}

The purpose of our paper is to develop a
principled technique for
attaching a probabilistic interpretation to feature
structures.
Our techniques apply to the
feature structures described by Carpenter~\cite{carpenter:ltfs}.
Since these structures are
the ones which are used in by
Pollard and Sag~\cite{pollard:sag:94} their relevance to
computational grammars is apparent.
On the basis of the usefulness of probabilistic context-free
grammars~\cite[ch. 5]{charniak:93},
it is plausible to assume that
 that the extension of probabilistic techniques to
such structures will allow the application
of known and new techniques of parse ranking and grammar
induction to more interesting grammars than has hitherto
been the case.

The paper is structured as follows. We start by reviewing
the training and use of probabilistic context-free grammars
(PCFGs).
We then develop a technique to allow
analogous probabilistic annotations on type hierarchies.
This gives us a clear account of the relationship between
a large class of feature structures and their probabilities,
but does not treat re-entrancy. We conclude by sketching
a technique which does treat such structures.
While
we know of previous work which associates scores
with feature structures~\cite{kim:94} are not aware
of any previous treatment which makes explicit the
link to classical probability theory.

We take a slightly unconventional perspective on feature structures,
because
it is easier to cast our theory within the more general framework of {\em
incremental description refinement}~\cite{mellish:88a}
than to exploit the usual metaphors of constraint-based grammar.
In fact we can afford to remain entirely agnostic about the
means by which the HPSG grammar associates signs with linguistic
strings, because all that we need in order to train our stochastic
procedures is a corpus of signs which are known to be valid
descriptions of strings.

\section{Probabilistic interpretation of PCFGs}

We review the standard probabilistic
interpretation of PCFGs~\footnote{Our description
is closely based on that given by Charniak\cite[p. 52 ff]{charniak:93}}.

A PCFG is a four-tuple
$< W, N, N_{1}, R >$ , where $W$ is a set of terminal symbols
$\lbrace w^{1},\ldots,w^{\omega}\rbrace$, $N$ is a set of non-terminal
symbols $\lbrace N^{1},\ldots,N^{\nu} \rbrace$, $N_{1}$ is the starting
symbol and $R$ is a set of rules of the form $N^{i} \rightarrow {\zeta}^{j}$,
where
${\zeta}^{j}$ is a string of terminals and non-terminals. Each rule
has a probability $P(N^{i} \rightarrow {\zeta}^{j})$ and the probabilities
for all the rules that expand a given non-terminal must sum to one.
We associate probabilities with {\em partial phrase markers}, which are sets
of terminal and non-terminal nodes generated by
beginning from the starting node successively
expanding non-terminal leaves of the partial tree. {\em Phrase markers}
are those partial phrase markers which have no non-terminal leaves.
Probabilities are assigned by the following inductive definition:
\begin{itemize}
	\item	$P(N_{1}) = 1$.
	\item  If $T$ is a partial phrase marker, and $T'$ is a partial
               phrase marker which differs from it only in that a single
               non-terminal node $N^k$ in $T$ has been expanded
	       to ${\zeta}^{m}$ in $T'$, then $P(T') = P(T) \times
	       P(N_k \rightarrow {\zeta}^{m})$.
\end{itemize}

In this definition $R$ acts as a specification of the {\em accessibility
relationships} which can hold between nodes of the trees admitted
by the grammar. The rule probabilities specify the cost of making
particular choices about the way in which the rules develop. It
is going to turn out that an exactly analogous
system of accessibility relations is present in the
probabilistic type hierarchies which we define later.

\paragraph{Limitations of PCFGs}
The definition of PCFGs implies that the probability of a phrase
marker depends only on the choice of rules used in expanding
non-terminal nodes. In particular, the probability does not depend on
the order in which the rules are applied.  This has the arguably
unwelcome consequence that PCFGs are unable to make certain
discriminations between trees which differ only in their
configuration~\footnote{The most obvious case is prepositional-phrase
attachment.}. The models developed in this paper
build in similar {\em independence assumptions}. A
large part of the art of probabilistic language modelling resides in
the management of the trade-off between descriptive power (which has
the merit of allowing us to make the discriminations which we want)
and independence assumptions (which have the merit of making training
practical by allowing us to treat similar situations as equivalent).

The crucial advantage of PCFGs over CFGs is that they can be trained
and/or learned from corpora. Readers for whom this
fact is unfamiliar are referred to Charniak's
textbook~\cite[Chapter 7]{charniak:93}.
We do not have space to recapitulate the
discussion of training which can be found there.
We do however illustrate the outcome of
training.

\subsection{Applying a PCFG to a simple corpus}

Consider the simple grammar in figure~\ref{lexicon}
\begin{figure*}
\begin{eqnarray*}
s  & \rightarrow & \mbox{\it np} \; \mbox{\it vp} \\
np & \rightarrow & \mbox{\it np-sing} \; | \; \mbox{\it np-pl} \\
vp  & \rightarrow & \mbox{\it vp-sing} \; | \;  \mbox{\it vp-pl}
\end{eqnarray*}
\begin{tabbing}
crossed \= \mbox{ \it vp-sing} \= crossed \= \mbox{ \it vp-sing} \kill
\\
bike \> \mbox{\it np-sing} \>
bus \> \mbox{\it np-sing} \\
car \> \mbox{\it np-sing} \>
cat \> \mbox{\it np-sing} \\
lorry \> \mbox{\it np-sing} \\
\\
bikes \> \mbox{\it np-pl} \>
buses \> \mbox{\it np-pl} \\
cars \> \mbox{\it np-pl} \>
cats \> \mbox{\it np-pl} \\
lorries \> \mbox{\it np-pl} \\
\\
stops \> \mbox{\it vp-sing} \>
crosses\> \mbox{\it vp-sing} \\
\\
stop\> \mbox{\it vp-pl} \>
cross\> \mbox{\it vp-pl}
\end{tabbing}

\caption{A simple grammar}
\label{lexicon}
\end{figure*}
and its training
against the corpus in figure~\ref{corpus}.
\begin{figure*}
\begin{tabbing}
{ trolleybus stopped} [x] \= { trolleybus stopped} [x] \= {cats stopped} \kill
{  car stops} \> {  bus stops} \>
{  lorries stop} \\
{  bikes stop} \> {  cats cross}
\end{tabbing}
\caption{A simple corpus}
\label{corpus}
\end{figure*}
Since there are 3 plural sentences
and only 2 singular sentences, the
optimal set of parameters will reflect the distribution found in the
corpus, as shown in figure~\ref{pcfg:training}
\begin{figure*}
\begin{eqnarray*}
P(\mbox{np}\; \mbox{vp} | s) &= & 1.0 \\
P(\mbox{np-sing} | np) & = & 0.45\\
P(\mbox{np-pl} | np) & = & 0.55 \\
P(\mbox{vp-sing} | vp) & = & 0.45 \\
P(\mbox{vp-pl} | vp) & = & 0.55 \\
\end{eqnarray*}
\caption{The results of training a PCFG}
\label{pcfg:training}
\end{figure*}
One might have hoped that the ratio
$P(\mbox{np-sing} | np)/P(\mbox{np-pl} | np)$
would be $2/3$, but it is instead $\sqrt{2/3}$. This is a
consequence of the assumption of independence. Effectively
the algorithm is ascribing the difference in distribution
of singular and plural sentences to the joint effect of
two independent decisions.
What we would really like it to do is to recognize that the
two apparently independent decisions are (in effect)
one and the same.
Also, because the grammar has no
means of enforcing number agreement, the system systematically
prefers plurals to singulars, even when doing this will lead
to agreement clashes. Thus ``buses stop''
has estimated $0.55 \times 0.55 = 0.3025$, ``bus stop'' and ``buses
stops'' both have probability $0.55 \times 0.45 = 0.2475$
and ``bus stops'' has probability $0.45 \times 0.45 = 0.2025$.
This behaviour is clearly unmotivated by the corpus, and arises
purely because of the inadequacy of the probabilistic model.

\section{Probabilistic type hierarchies}

\paragraph{ALE signatures}
Carpenter's ALE~\cite{carpenter:ale}
allows  the user to define the {\em type hierarchy} of a grammar by
writing a collection
of clauses which together denote an {\em inheritance hierarchy},
a set of {\em features} and a set of {\em appropriateness conditions}.
An example of such a hierarchy is given in
ALE syntax in figure~\ref{hierarchy}.
\begin{figure*}
\begin{verbatim}
bot sub [sign,num].
    sign sub [sentence,phrase].
     sentence sub []
              intro [left:np,right:vp].
     phrase sub [np,vp]
           intro [num:num].
      np sub [].
      vp sub [].
     num sub [sing,pl].
      sing sub [].
      pl sub [].
\end{verbatim}
\caption{An ALE signature}
\label{hierarchy}
\end{figure*}

\paragraph{What the ALE signature tells us}
The inheritance information tells us that a sign is a
forced choice between a sentence and a phrase, that a phrase is a forced
choice between a noun-phrase (\verb+np+) and a verb-phrase
(\verb+vp+) and that number values (\verb+num+) are partitioned into
singular (\verb+sing+) and plural (\verb+pl+).
The features which are
defined are \verb+left+,\verb+right+, and \verb+num+, and the appropriateness
information says that the feature \verb+num+ {\em introduces} a new
instance of the type \verb+num+ on all \verb+phrases+, and that \verb+left+ and
\verb+right+ introduce \verb+np+ and \verb+vp+ respectively
on \verb+sentences+.

\paragraph{The parallel with PCFGs}
The parallel which makes it possible to apply the PCFG training
scheme almost unchanged is that the sub-types of a given super-type
partition the feature structures of that type in just the same
way that the different rules which expand a given non-terminal
$N$ of the PCFG partition the space
of trees whose topmost node is $N$. Equally, the features defined
in the hierarchy act as an accessibility relation between nodes in
a way which is for our purposes entirely equivalent to the way
in which the right hand sides of the rules introduce new nodes
into partial phrase markers~\footnote{Each rule of a PCFG also
specifies a total ordering over the nodes which it introduces, but
the training algorithm does not rely on this fact}. The hierarchy
in figure~\ref{hierarchy} is related to but not isomorphic with
the grammar in figure~\ref{lexicon}.

One difference is that \verb+num+
is explicitly introduced as a feature in the hierarchy, where at is
only implicitly present in the original grammar. The other difference
is the use of \verb+left+ and \verb+right+ as models
of the dominance relationships between nodes.

\section{A probabilistic interpretation of  typed feature-structures}

For our purposes, a probabilistic type hierarchy (PTH) is a four-tuple
\[< MT, NT, NT_{1}, I >\]
where $MT$ is a set of maximal
types~\footnote{We follow Carpenter's convention for types.
The bottom node is the
one containing no information, and the maximal nodes are the ones
containing the maximum amounts of information possible.
}
$\lbrace t^{1},\ldots,t^{\omega}\rbrace$, $NT$ is a set of non-maximal
types $\lbrace T^{1},\ldots,T^{\nu} \rbrace$, $NT_{1}$ is the starting
symbol and $I$ is a set of
introduction relationships
of the form $(T^{i} \Rightarrow T^{j}) \rightarrow {\xi}^{k}$, where
${\xi}^{j}$ is a multiset of maximal and non-maximal types. Each introduction
relationship
has a probability $P((T^{i} \Rightarrow T^{j}) \rightarrow {\xi}^{k})$ and the
probabilities
for all the introduction relationships that apply to
a given non-maximal type must sum to one.

As things stand this definition
is nearly isomorphic
to that given for PCFGs, with the major differences
being two changes which move us from rules to introduction relationships.
Firstly, we relax the stipulation that the items on the right hand side
of the rules are strings, allowing them instead to be multisets.
Secondly, we introduce an additional term in the head of
introduction rules to signal the fact that when we apply a
particular introduction relationship to a node we also
specialize the type of the node  by picking exactly one
of the direct subtypes of its current type. Finally, we need to deal with the
case where $T^{j}$ is non-maximal. This is simply achieved by defining the
{\em iterated introduction relationships} from $T^{i}$ as being those
corresponding to the chains of introduction relationships from $T^{i}$ which
refine the type to a maximal type.
In the probabilistic type hierarchy, it is the iterated introduction
relationships which correspond to the context-free rewrite rules of
a PCFG. A useful side-effect of this is that we can preserve the
invariant that all types except those at the fringe of the structure
are maximal.

The hierarchy whose ALE syntax is given in figure~\ref{hierarchy}
is captured in the new notation by figure~\ref{formal:hierarchy}
\begin{figure*}
\begin{eqnarray*}
MT  & = & \lbrace \mbox{sentence},\mbox{np},\mbox{vp},\mbox{sing},\mbox{pl}
\rbrace \\
NT & = & \lbrace \mbox{bot},\mbox{sign},\mbox{phrase},\mbox{num} \rbrace \\
NT_{1} & = & \mbox{bot} \\
I   & = & \lbrace (\mbox{bot} \Rightarrow \mbox{sign}) \rightarrow [] \\
	& & (\mbox{bot} \Rightarrow \mbox{num}) \rightarrow [] \\
    & & (\mbox{sign} \Rightarrow \mbox{sentence}) \rightarrow
[\mbox{np},\mbox{vp}] \\
    & &  (\mbox{sign} \Rightarrow \mbox{phrase}) \rightarrow [\mbox{num}] \\
    & &     (\mbox{phrase} \Rightarrow \mbox{np}) \rightarrow [] \\
    & &     (\mbox{phrase} \Rightarrow \mbox{vp}) \rightarrow [] \\
    & &     (\mbox{num} \Rightarrow \mbox{sing}) \rightarrow [] \\
    & &     (\mbox{num} \Rightarrow \mbox{pl}) \rightarrow [] \rbrace
\end{eqnarray*}
\caption{A more formal version of the simple hierarchy}
\label{formal:hierarchy}
\end{figure*}

We associate probabilities with feature structures, which are sets
of maximal  and non-maximal nodes generated by
beginning from the starting node and successively
expanding non-maximal leaves of the partial tree.
{\em Maximally
specified feature structures}
are those feature structures which have only maximal leaves.
Probabilities are assigned by the following inductive definition:
\begin{itemize}
	\item	$P(NT_{1}) = 1$.
	\item  If $F$ is a feature structure, and $F'$ is a partial
               feature structure which differs from it only in that a single
               non-maximal node $NT^k$ of type ${T_{0}}^k$
               in $F$ has been refined to type ${T_{1}}^k$ expanded
	       to ${\xi}^{m}$ in $F'$, then $P(F') = P(F) \times
	       P((T0 \Rightarrow T1) \rightarrow {\xi}^{m})$.
\end{itemize}

Modulo notation, this definition is identical to the one
given earlier for PCFGs.
Given the correspondence between the definitions of a PTH and
a PCFG it should be apparent that the training methods
which apply to one can equally be used with the
other. We will shortly provide an example.
Because we have not yet treated the crucial matter of re-entrancy,
it would be inappropriate to call what we so far have stochastic HPSG,
so we refer to it as stochastic $\mbox{HPSG}^-$.

\subsection{Using stochastic $\mbox{HPSG}^-$ with the corpus}
Using the hierarchy in figure~\ref{hierarchy}
the analyses of the five sentences
{} from figure~\ref{corpus} are as in figure~\ref{analyses}.

\begin{figure*}
\begin{displaymath}
\typedavm{vp} {\attval{LEFT}{\typedavm{np}{\attval{NUM}{sing}}}
                     \attval{RIGHT}{\typedavm{vp}{\attval{NUM}{sing}}}}
\end{displaymath}
(2 occurrences)
\begin{displaymath}
\typedavm{vp} {\attval{LEFT}{\typedavm{np}{\attval{NUM}{pl}}}
                     \attval{RIGHT}{\typedavm{vp}{\attval{NUM}{pl}}}}
\end{displaymath}
(3 occurrences).
\caption{Analyses of the corpus using the ALE-hierarchy}
\label{analyses}
\end{figure*}

Training is a matter of counting the transitions which
are found the observed results, then using
counts to refine initial estimates of the probabilities
of particular transitions. This is entirely analogous to
what went on with PCFGs.
The results of training are essentially identical to those given earlier,
with the optimal assignment being as shown in figure~\ref{pth:training}.
At this point we have provided a system which allows us to use
feature structures instead of PCFGs, but we have not yet
dealt with the question of re-entrancy, which forms
a crucial part of the expressive power of typed feature
structures. We will return to this shortly, but first
we consider the detailed implications of what we have done so
far.
\begin{figure*}
\begin{eqnarray*}
P(\mbox{bot} \Rightarrow \mbox{sign}) & = & 1.0\\
P(\mbox{bot} \Rightarrow \mbox{num}) & = & 0.0 \\
P(\mbox{sign} \Rightarrow \mbox{sentence}) & = & 1.0\\
P(\mbox{sign} \Rightarrow \mbox{phrase}) & = & 0.0 \\
P(\mbox{num} \Rightarrow \mbox{sing}) & = & 0.45 \\
P(\mbox{num} \Rightarrow \mbox{pl}) & = & 0.55 \\
\\
P(\mbox{phrase} \Rightarrow \mbox{np}) & = & \lambda \\
P(\mbox{phrase} \Rightarrow \mbox{vp}) & = & 1-\lambda
\end{eqnarray*}
\caption{The results of training the probabilistic type hierarchy}
\label{pth:training}
\end{figure*}
The similarities between these results and those in figure~\ref{pcfg:training}
\begin{itemize}
\item We still model the distribution observed in the corpus
by assuming two independent decisions.
\item We still get a strange ranking of the parses, which favours
number disagreement,in spite
of the fact that the grammar which generated the corpus enforces
number agreement.
\end{itemize}
The differences between these results and the earlier ones
are:
\begin{itemize}
\item The hierarchy uses \verb+bot+ rather than \verb+s+ as its
      start symbol. The probabilities tell us that the
      corpus contains no free-standing structures of type \mbox{num}.
\item The zero probability of
\[ \mbox{sign} \Rightarrow \mbox{phrase} \]
      codifies a similar observation that there are no free-standing
      structures with type \verb+phrase+.

\item Since items of type \mbox{phrase} are never introduced at that
      type, but only in the form of sub-types,
      there are no transitions from \verb+phrase+ in the corpus.
      Therefore the initial estimates of the probabilities of such transitions
      are unaffected by training.
\item In the PCFG the symmetry between the expansions of \verb+np+
      and \mbox{vp} to singular and plural variants is implicit, whereas
      in the PTH the distribution of singular and plural variants
      is encoded at a single location, namely that at which \verb+num+
      is refined.
\end{itemize}

The independence assumption which is built into the
training algorithm is that types are to be refined
according to the same probability distribution
irrespective of the context in which they are expanded.
We have already seen a consequence of this: the PTH
lumps together all occasions where \mbox{num} is
expanded, irrespective of whether the enclosing context
is \mbox{np} or \mbox{vp}.
For the moment we
are prepared to tolerate this because:
\begin{itemize}
\item {\bf Clarity:} The decisions which we have made lead to a
       system with a clear probabilistic semantics.
\item {\bf Trainability: } the number of parameters which must be
      estimated for a grammar is a linear function of the size of the
      type hierarchy
\item {\bf Easy extensibility: } There is a clear route to a more
      finely grained account if we allow the expansion probabilities
      to be conditioned on surrounding context. This would
      increase the number of parameters to be estimated, which
      may or may not prove to be a problem.
\end{itemize}

\section{Adding re-entrancies}

We now turn to an extension of the system which takes proper
account of re-entrancies in the structure. The essence of our
approach is to define a stochastic procedure which simultaneously
expands the nodes of the tree in the way outlined above and
guesses the pattern of re-entrancies which relate them. It
pays to stipulate that the structures which we build are {\em fully
inequated} in the sense defined by Carpenter~\cite[p120]{carpenter:ltfs}.

The essential insight is that the choice of a fully inequated feature
structure involving a set of nodes is the same thing as the choice of
an arbitrary equivalence relation over these nodes, and this is in
turn equivalent to the choice of a partition of the set of nodes into
a set of non-empty sets. These sets of nodes are equivalence classes.
The standard recursive procedure for generating partitions of $k+1$
elements is to non-deterministically add the $k+1$thq node to each of
the equivalence classes of each of the partitions of $k$ nodes, and
also to nondeterministically consider the new node as a singleton set.
The basis of the stochastic procedure for generating fully-inequated
feature structures is to interleave the generation of equivalence
classes with the expansion from the initial node as described above.

For the purposes of the expansion algorithm, a fully inequated feature
structure consists of a feature tree (as before) and an equivalence
relation\footnote{Since maximal types are mutually inconsistent, this
equivalence relation can be efficiently represented by a associating a
separate partition with each maximal type} over all the maximal
nodes in that tree. The task of the algorithm is to generate all such
structures and to equip them with probabilities.  We proceed as in the
case without re-entrancy, except that we only ever expand sub-trees in
the case where the new node begins a new equivalence class.  This
avoids the double counting which was a problem earlier.

The remaining task is that of assigning scores to equivalence
relations. We do not have a fully satisfactory solution to this
problem. The reason for this is that we would ideally like to assign
probabilities to intermediate structures in such a way that the
probabilities of fully expanded structures are independent of the
route by which they were arrived at. This can be done, and the method
which we adopt has the merit of simplicity.

\subsection{Scoring re-entrancies}

We associate a single
probabilistic parameter $P(T_{=})$ with each type $T$, and derive the
probability of the structure in which a particular pairwise equation
of nodes in type $T$ have been equated by multiplying the probability
of the structure in which no decision has been made by $P(T_{=})$. We
derive the probability of the corresponding inequated structure by
multiplying by $1-P(T_{=})$ in an entirely analogous way. This ensures
that the probabilities of the equated and inequated extensions of
the original structure sum to the original probability. The cost is
a deficiency in modelling, since this takes no account of the fact
that token identity of nodes is transitive.
which are generated. As things stand the stochastic
procedure is free to generate structures where $n_{1} \equiv n_{2}$,
$n_{2} \equiv n_{3}$ but $n_{1} \not\equiv n_{3}$, which are not in fact
legal feature structures.
This leads to distortions of the probability estimates since the training
algorithm spends part of its probability mass on impossible structures.

\subsection{Evaluation}

Even a crude account of re-entrancy is better than completely ignoring
the issue, and the one proposed gets the right result for cases of
double counting such as those discussed above, but it should be
obvious that there is room for improvement in the treatment which we
provide. Intuitively what is required is a parametrisable means of
distributing probability mass among the distinct equivalence relations
which extend the current structure. One attractive possibility
would be to enumerate the relations which can be obtained by adding
the current node to the various different equivalence
classes which are available, apply some scoring function to each
class, and then normalize such that the total score over all alternatives
is one.
But this might
introduce unpleasant dependencies of the probabilities of feature
structures on the order in which the stochastic procedure chooses to
expand nodes,
because the normalisation is carried out before we have full knowledge
of the equivalence classes with which the current node might become
associated.
It may be that an appropriate choice of scoring function will
circumvent this difficulty, but this is left as a matter for
further research.

\section{Conclusions}

We have presented two proposals for the association of probabilities
with typed feature-structures of the form used in HPSG. As far as we
know these are the most detailed of their type, and the ones which are
most likely to be able to exploit standard training and parsing
algorithms. For typed feature structures lacking re-entrancy we
believe our proposal to be the simplest and most natural which is
available.
The proposal for dealing with re-entrancy is less satisfactory
but offers a basis for empirical exploration, and has definite
advantages over the straightforward use of PCFGs.
We plan to follow up the current work by  training and testing
a suitable instantiation of our framework  against
manually annotated corpora.

\section{Acknowledgements}

I acknowledge the support of the Language Technology Group of the
Human Communication Research Centre, which is a UK ESRC
 funded
institution.


\begin{thebibliography}{}

\bibitem[\protect\citename{Carpenter}1992]{carpenter:ltfs}
Bob Carpenter.
\newblock 1992.
\newblock {\em The Logic of Typed Feature Structures}.
\newblock Cambridge Tracts in Theoretical Computer Science. CUP.
\newblock With Applications to Unification Grammars, Logic Programs and
  Constraint Resolution.

\bibitem[\protect\citename{Carpenter}1993]{carpenter:ale}
Bob Carpenter, 1993.
\newblock {\em ALE. The Attribute Logic Engine user's guide, version $\beta$.}
\newblock Carnegie Mellon University, Pittsburgh, Pa., Laboratory for
  Computational Linguistics, MAY.

\bibitem[\protect\citename{Charniak}1993]{charniak:93}
Eugene Charniak.
\newblock 1993.
\newblock {\em Statistical Language Learning}.
\newblock The MIT Press.

\bibitem[\protect\citename{Kim}1994]{kim:94}
Albert Kim.
\newblock 1994.
\newblock Graded unification: A framework for interactive processing.
\newblock In {\em Proceedings of the 32nd Annual Meeting of the Association for
  Computational Linguistics}, pages 313--315, June.

\bibitem[\protect\citename{Mellish}1988]{mellish:88a}
C.S. Mellish.
\newblock 1988.
\newblock Implementing systemic classification by unification.
\newblock {\em Computational Linguistics}, 14(1):40--51.
\newblock Winter.

\bibitem[\protect\citename{Pollard and Sag}1994]{pollard:sag:94}
Carl Pollard and Ivan~A. Sag.
\newblock 1994.
\newblock {\em Head-Driven Phrase Structure Grammar}.
\newblock CSLI and University of Chicago Press, Stanford, Ca.\ and Chicago,
  Ill.

\end{thebibliography}
\end{document}